\documentclass[10pt]{article}
\usepackage[left=2.54cm,right=2.54cm,top=2.54cm,bottom=2.54cm]{geometry}
\usepackage{cite}
\usepackage{amsmath,amssymb,amsfonts}
\usepackage{algorithmic}
\usepackage{graphicx}
\usepackage{textcomp}
\usepackage{xcolor}
\def\BibTeX{{\rm B\kern-.05em{\sc i\kern-.025em b}\kern-.08em
    T\kern-.1667em\lower.7ex\hbox{E}\kern-.125emX}}
\usepackage{graphicx}

\usepackage{pgfplots}
\usepackage{pgfplotstable}
\usepackage{booktabs, colortbl, array}
\usepackage{enumitem}
\usepackage{rotating}
\usepackage{balance}
\usepackage{bbm}
\usepackage{amssymb}
\usepackage{amsmath}
\usepackage{amsthm}

\usepackage{url}
\newtheorem{definition}{Definition}

\begin{document}
\title{Mathematical Models for Local Sensing Hashes}
%
%
\author{Lilon Wangner}

\maketitle              
\begin{abstract}
As data volumes continue to grow, searches in data are becoming increasingly time-consuming. Classical index structures for neighbor search are no longer sustainable due to the ``curse of dimensionality''. Instead, approximated index structures offer a good opportunity to significantly accelerate the neighbor search for clustering and outlier detection and to have the lowest possible error rate in the results of the algorithms. Local sensing hashes is one of those. We indicate directions to mathematically model the properties of it.
\end{abstract}

\section{Basic Definition}
In a local sensing hashes we use a number $m$ of alternative clusterings, and we refine for each of the $m$ clusterings exactly one cell, in which the query point (origin) is located. Therefore, the selectivity of the query response corresponds to the intersection of the union of all $m$ cells with the query ball. We start with the analysis of the case $s=b=1$. We assume that each of the $m$ cells is selected uniformly and independently such that the origin (query point) is located inside, so the upper boundary of each cell in each dimension is uniformly taken from the interval $[0..\tfrac{1}{2}]$ and the lower boundary is consequently from $[-\tfrac{1}{2}..0]$.

The idea is to consider the convex, Voronoi-like cells of k-means as a $d$-dimensional grid of hyper-cubic cells which have a volume such that an expectation of $n/k$ points are located within each. We denote the side length of these cells with $b$. The query is modeled as a ball with perimeter $s$, centered at origin. In the beginning, we assume that the ball is actually from a maximum metric such that the query is as well a hypercube with side length $s$.

In the following  we will use several definite integrals which we denote in a bit unusual way which helps for clarity we will write
\[\int ... \mbox{ }(0\le x \le 1) \mbox{ instead of }\int_0^1 ... \mbox{ d}x.\]
We will need a few integrals throughout this paper:
\[\int x+\tfrac{1}{2}\mbox{ }(0\le x \le\tfrac{1}{2})  =  \tfrac{3}{8}\]
\[\int\hspace{-3mm}\int (x+\tfrac{1}{2})(y+\tfrac{1}{2})\mbox{ }(0\le x,y \le\tfrac{1}{2})  =  \tfrac{9}{64}\]
\[\int...\int (x_1+\tfrac{1}{2})\cdot ...\cdot (x_d+\tfrac{1}{2})\mbox{ }(0\le x_1, ..., x_d \le\tfrac{1}{2})  =  \Big(\tfrac{3}{8}\Big)^d\]
\[\int\hspace{-3mm}\int\min(x_1,x_2)+\tfrac{1}{2}\mbox{ }(0\le x_1, x_2 \le\tfrac{1}{2})  =  \tfrac{5}{24}\]
\begin{eqnarray*}\int...\int\min(x_1,...,x_d)+\tfrac{1}{2}\mbox{ }(0\le x_1, ... , x_d \le\tfrac{1}{2}) & = \\
=d\cdot \int x^{d-1}\cdot\big(x+\tfrac{1}{2}\big) \mbox{ }(0\le x \le\tfrac{1}{2})& = & \frac{2d+1}{(d+1)\cdot 2^{d+1}}
\end{eqnarray*}
\[\int\hspace{-3mm}\int y-v \mbox{ }(0\le y \le\tfrac{1}{2}, -\tfrac{1}{2}\le v \le 0) = \tfrac{1}{8}\]
\[\int ... \int (y_1-v_1)\cdot ... \cdot (y_d-v_d) \mbox{ }(0\le y_1, ..., y_d \le\tfrac{1}{2}, -\tfrac{1}{2}\le v_1, ..., v_d \le 0) = \Big(\tfrac{1}{8}\Big)^d\]
And, finally, we can also solve the following combination:
\[\int...\int \big(\min(x_1, ..., x_d)+\tfrac{1}{2}\big)\cdot(y_1-v_1)\cdot ... \cdot (y_m-v_m) \mbox{ }(0\le x_1, ..., x_d, y_1, ..., y_m \le\tfrac{1}{2},-\tfrac{1}{2}\le v_1, ..., v_m \le 0)\]
which gives:
\[=\frac{2m+1}{8^d\cdot(m+1)\cdot 2^{m+1}}.\]
\begin{definition}
$p(m,\ell,d)$ is the (hyper-) volume of the hypercube representing the query which is occupied by at least $\ell$ of the $m$ cells (with $\ell\le m$). Here the $m$ cells are uniformly selected in $\mathbb R^d$ such that the query point is inside the cell.
\end{definition}
Note that $p(m,1,d)$ corresponds to the selectivity of the hashing provided that $b$ and $s$ are of equal size.

For the case $m=\ell=1$ we can easily derive the closed formula of $p(1,1,d)$. We simply have to form, in each quadrant, the (identical) expectation with which the query is occupied by the cell:
\[p(1,1,d) = 2^d\cdot \int...\int (x_1+\tfrac{1}{2})\cdot ... \cdot (x_d+\tfrac{1}{2})\mbox{ }(0\le x_1, ..., x_d \le\tfrac{1}{2})  =  \Big(\tfrac{3}{4}\Big)^d\]
The case $m>1, \ell=1$ can be reduced to the case $m=1$:
\[p(m,1,d) = \bigg({m\atop 1}\bigg)\cdot p(1,1,d) - \bigg({m\atop 2}\bigg)\cdot p(1,2,d) + \bigg({m\atop 3}\bigg) \cdot p(1, 3, d) - ... \pm \bigg({m\atop m}\bigg)\cdot p(1, m-1, d)\]

For the case $m=1, \ell=2$, we start with the analysis of $d=1$ and construct this case with two variables $x$ and $y$. We want to estimate the area which is covered by $x$ and $y$, which is
\[p(1,2,1) = \int\hspace{-3mm}\int\left\{\begin{array}{ll}1-|x-y|&\mbox{ if }x<0\mbox{ xor } y<0\\
1-\max(|x|,|y|)&\mbox{ otherwise}\end{array}\right.(-\tfrac{1}{2}\le x,y \le\tfrac{1}{2})=\frac{7}{12}.\]
This is an independent information in each dimension such that
\[p(1,2,d) = p(1,2,1)^d = \Big(\frac{7}{12}\Big)^d.\]
For a generalization to $\ell=3$ we make a similar case distinction:
\[p(1,3,1)=\int\hspace{-3mm}\int\hspace{-3mm}\int \left\{\begin{array}{ll}
    1-\max(x,y,z)&\mbox{if }x,y,z\ge 0\\
    1-\max(x,y)+z&\mbox{if }x,y\ge 0\mbox{ and } z< 0\\
    ...
\end{array}\right.(...) = \frac{15}{32}.\]
where the first case occurs in 2 octants and the second case in the remaining 6 octants. We solve the first case by letting
\[2\cdot \int\hspace{-3mm}\int\hspace{-3mm}\int 1-\max(x,y,z) = 6\cdot \int (1-x)\cdot x^2 = \tfrac{5}{32}\]
and similarly the second case
\[6\cdot \int\hspace{-3mm}\int\hspace{-3mm}\int 1-\max(x,y)+z = 12\cdot \int\hspace{-3mm}\int (1-x +z)\cdot x =\tfrac{5}{16}.\]

For the general case, we consider $\ell$ variables $x_1, ..., x_\ell$. Again we consider a solution space where we distinguish if each variable is greater or less than 0. All variables are equivalent, thus we consider all quadrants of the solutions space equally where the same number of variables is $<0$. There is a number of\[\left({\ell \atop i}\right)\mbox{ quadrants of the solution space having $i$ out of $\ell$ variables} <0.\]
Therefore, we have
\[p(1,\ell,1) = \sum_{0\le i\le \ell}\left({\ell \atop i}\right)\cdot \int...\int 1-\max(x_1, ..., x_i)+\min(x_{i+1}, ..., x_\ell)\hspace{5mm}(0\le x_1, ..., x_i\le \tfrac{1}{2}, -\tfrac{1}{2}\le x_{i+1}, ..., x_\ell\le 0)\]

\section{Related Work}

\textbf{Approximate} nearest neighbor search techniques can also be applied to the similarity join problem, however without guarantees on completeness and exactness of the result. There may be false positives as well as false negatives. Recently an approach \cite{DBLP:journals/tkde/YuNLWY17} to Local Sens Hash is used on a representative point sample, to reduce the number of lookup operations. LSH is of interest in theoretical foundational work, where a recursive and cache-oblivous LSH approach \cite{DBLP:journals/algorithmica/PaghPSS17} was proposed. The topic of approximate solutions for the similarity join is also an emerging field in deep learning \cite{DBLP:journals/corr/abs-1803-04765}. There are approximative approaches which target low dimensional cases (spatial joins in 2--3 dimensions \cite{DBLP:conf/icde/BryanEF08}) or higher (10--20) dimensional cases  \cite{DBLP:conf/focs/AndoniI06}. Very high-dimensional cases, with dimensions of $128$ and above have been targeted with Symbolic Aggregate approXimation (SAX) techniques \cite{DBLP:journals/concurrency/MaJZ17}) to generate approximate candidates. SAX techniques rely on several indirect parameters like PAA size or the iSAX alphabet size.
\nocite{DBLP:conf/kdd/DittrichS01, DBLP:journals/tkde/KoudasS00, DBLP:conf/icde/LiebermanSS08} \nocite{DBLP:journals/tkde/ChenGLJC17}.
\nocite{DBLP:conf/icde/LiebermanSS08} \nocite{DBLP:journals/tkde/KoudasS00}  \nocite{DBLP:journals/tkde/ChenGLJC17}\nocite{epsilongridorder}\nocite{DBLP:conf/dasfaa/KalashnikovP03} \nocite{DBLP:journals/is/KalashnikovP07}\nocite{DBLP:journals/vldb/Kalashnikov13}  \nocite{DBLP:journals/vldb/Kalashnikov13} \nocite{DBLP:conf/sigmod/SilvaR12, DBLP:conf/cloudi/SilvaRT12}\nocite{DBLP:conf/icde/BohmK01, DBLP:journals/jda/ParedesR09, DBLP:journals/tkde/ChenGLJC17}\nocite{DBLP:conf/sigmod/BrinkhoffKS93}\nocite{DBLP:conf/vldb/CiacciaPZ97}\nocite{DBLP:journals/jda/ParedesR09}, \nocite{DBLP:journals/mta/DohnalGSZ03} \nocite{DBLP:conf/dexa/DohnalGZ03} \nocite{DBLP:conf/sisap/PearsonS14}) \nocite{DBLP:conf/stoc/IndykM98}.

\nocite{DBLP:conf/sigmod/SchneiderD89, DBLP:conf/kdd/WangMP13, DBLP:journals/pvldb/FierABLF18}\nocite{DBLP:conf/waim/LiWU16, DBLP:conf/sigmod/McCauley018, DBLP:journals/pvldb/FierABLF18} \nocite{DBLP:conf/sigmod/ZhaoRDW16}\nocite{DBLP:conf/btw/BohmNPZ09}\nocite{DBLP:conf/icde/LiebermanSS08}\nocite{DBLP:conf/icde/LiebermanSS08}
\nocite{DBLP:conf/btw/BohmNPZ09}
\nocite{DBLP:conf/focs/FrigoLPR99}\nocite{DBLP:conf/sigmod/HeLLY07} \nocite{DBLP:journals/tog/MoonBKCKBNY10}\nocite{DBLP:journals/bmcbi/FerreiraRR14}\nocite{DBLP:conf/para/BaderM06, DBLP:conf/europar/Bader08}\nocite{loopsjournal}\nocite{DBLP:conf/bigdataconf/BohmPP16, IEEE:transbigdata/BohmPP18}\nocite{DBLP:journals/toms/DongarraCHD90} 

\section{Conclusion}
As data volumes continue to grow, searches in data are becoming increasingly time-consuming. Classical index structures for neighbor search are no longer sustainable due to the ``curse of dimensionality''. Instead, approximated index structures offer a good opportunity to significantly accelerate the neighbor search for clustering and outlier detection and to have the lowest possible error rate in the results of the algorithms. Local sensing hashes is one of those. We indicate directions to mathematically model the properties of it.
There is still a lot research necessary with local sensing hashes.

\nocite{DBLP:conf/sigmod/BohmFP08,DBLP:conf/icdt/BerchtoldBKK01,DBLP:conf/icdm/BohmK02,10.1007/BFb0000120,DBLP:conf/icde/BohmOPY07,
DBLP:conf/cikm/BohmBBK00,DBLP:journals/jiis/BohmBKM00,DBLP:conf/adl/BohmBKS00,DBLP:conf/edbt/BohmK00,
DBLP:journals/sadm/AchtertBDKZ08,DBLP:journals/bioinformatics/BaumgartnerBBMWOLR04,DBLP:journals/jbi/BaumgartnerBB05,
DBLP:conf/icdt/BerchtoldBKK01,DBLP:conf/edbt/BohmP08,DBLP:conf/kdd/AchtertBKKZ06,DBLP:conf/cikm/BohmFOPW09,
DBLP:conf/ssdbm/BohmPS06,DBLP:conf/kdd/BohmHMP09,DBLP:conf/dawak/BerchtoldBKKX00,DBLP:conf/sdm/AchtertBDKZ08,
DBLP:journals/jdi/BaumgartnerGBF05,DBLP:journals/kais/MaiHFPB15,DBLP:journals/tlsdkcs/BohmNPWZ09,
DBLP:conf/dexa/BohmK03,DBLP:conf/icde/BohmGKPS07,DBLP:conf/miccai/DyrbaEWKPOMPBFFHKHKT12,DBLP:conf/kdd/PlantB11,
DBLP:journals/bioinformatics/PlantBTB06,DBLP:conf/icdm/ShaoPYB11,DBLP:conf/icdm/PlantWZ09,alzheimer,
DBLP:journals/tkdd/BohmFPP07,DBLP:conf/pakdd/BohmGOPPW10,DBLP:conf/btw/BohmNPZ09,DBLP:conf/icdm/MaiGP12,
DBLP:journals/kais/ShaoWYPB17,DBLP:conf/kdd/YeGPB16,DBLP:conf/kdd/FengHKBP12,DBLP:journals/envsoft/YangSSBP12,
DBLP:conf/icdm/GoeblHPB14,DBLP:conf/kdd/AltinigneliPB13,DBLP:conf/icdm/YeMHP16,DBLP:conf/kdd/Plant12,
DBLP:conf/kdd/Plant12,DBLP:conf/cikm/BohmBBK00,DBLP:journals/kais/BohmK04,DBLP:conf/icdm/BohmK02,DBLP:conf/icdm/BohmK02, DBLP:conf/mlg/WackersreutherW10,DBLP:conf/ssdbm/AchtertBKKZ07,DBLP:journals/tkde/ShaoHBYP13,DBLP:conf/pkdd/AchtertBKKMZ06,DBLP:conf/ssdbm/AchtertBKZ06,
patent,DBLP:conf/kdd/MaurusP16,DBLP:conf/icdm/MaiHHPB13,water,DBLP:journals/isci/ShaoTGYPA19,DBLP:conf/sdm/BohmPP17,DBLP:conf/icdm/GoeblTBP16,DBLP:conf/kdd/MautzYPB17,
DBLP:conf/icdt/BerchtoldBKK01,loopsjournal,Bially1969SpacefillingCT,
Prusinkiewicz:1986:GAL:16564.16608,10.1007/BFb0000120}
\bibliographystyle{unsrt}
\bibliography{bibliography}

\begin{thebibliography}{10}

\bibitem{DBLP:journals/tkde/YuNLWY17}
Chenyun Yu, Sarana Nutanong, Hangyu Li, Cong Wang, and Xingliang Yuan.
\newblock A generic method for accelerating lsh-based similarity join
  processing.
\newblock {\em {IEEE} Trans. Knowl. Data Eng.}, 29(4):712--726, 2017.

\bibitem{DBLP:journals/algorithmica/PaghPSS17}
Rasmus Pagh, Ninh Pham, Francesco Silvestri, and Morten St{\"{o}}ckel.
\newblock I/o-efficient similarity join.
\newblock {\em Algorithmica}, 78(4):1263--1283, 2017.

\bibitem{DBLP:journals/corr/abs-1803-04765}
Nicolas Papernot and Patrick~D. McDaniel.
\newblock Deep k-nearest neighbors: Towards confident, interpretable and robust
  deep learning.
\newblock {\em CoRR}, abs/1803.04765, 2018.

\bibitem{DBLP:conf/icde/BryanEF08}
Brent Bryan, Frederick Eberhardt, and Christos Faloutsos.
\newblock Compact similarity joins.
\newblock In {\em {ICDE}}, pages 346--355, 2008.

\bibitem{DBLP:conf/focs/AndoniI06}
Alexandr Andoni and Piotr Indyk.
\newblock Near-optimal hashing algorithms for approximate nearest neighbor in
  high dimensions.
\newblock In {\em FOCS 2006}, pages 459--468, 2006.

\bibitem{DBLP:journals/concurrency/MaJZ17}
Youzhong Ma, Shijie Jia, and Yongxin Zhang.
\newblock A novel approach for high-dimensional vector similarity join query.
\newblock {\em Concurrency and Computation: Practice and Experience}, 29(5),
  2017.

\bibitem{DBLP:conf/kdd/DittrichS01}
Jens{-}Peter Dittrich and Bernhard Seeger.
\newblock {GESS:} a scalable similarity-join algorithm for mining large data
  sets in high dimensional spaces.
\newblock In {\em {SIGKDD}}, pages 47--56, 2001.

\bibitem{DBLP:journals/tkde/KoudasS00}
Nick Koudas and Kenneth~C. Sevcik.
\newblock High dimensional similarity joins: Algorithms and performance
  evaluation.
\newblock {\em {IEEE} Trans. Knowl. Data Eng.}, 12(1):3--18, 2000.

\bibitem{DBLP:conf/icde/LiebermanSS08}
Michael~D. Lieberman, Jagan Sankaranarayanan, and Hanan Samet.
\newblock A fast similarity join algorithm using graphics processing units.
\newblock In {\em {ICDE}}, pages 1111--1120, 2008.

\bibitem{DBLP:journals/tkde/ChenGLJC17}
Lu~Chen, Yunjun Gao, Xinhan Li, Christian~S. Jensen, and Gang Chen.
\newblock Efficient metric indexing for similarity search and similarity joins.
\newblock {\em {IEEE} Trans. Knowl. Data Eng.}, 29(3):556--571, 2017.

\bibitem{epsilongridorder}
Christian B{\"{o}}hm, Bernhard Braunm{\"{u}}ller, Florian Krebs, and Hans-Peter
  Kriegel.
\newblock Epsilon grid order: An algorithm for the similarity join on massive
  high-dimensional data.
\newblock In {\em SIGMOD Conf. 2001}, pages 379--388, 2001.

\bibitem{DBLP:conf/dasfaa/KalashnikovP03}
Dmitri~V. Kalashnikov and Sunil Prabhakar.
\newblock Similarity join for low-and high-dimensional data.
\newblock In {\em {(DASFAA} '03)}, pages 7--16, 2003.

\bibitem{DBLP:journals/is/KalashnikovP07}
Dmitri~V. Kalashnikov and Sunil Prabhakar.
\newblock Fast similarity join for multi-dimensional data.
\newblock {\em Inf. Syst.}, 32(1):160--177, 2007.

\bibitem{DBLP:journals/vldb/Kalashnikov13}
Dmitri~V. Kalashnikov.
\newblock Super-ego: fast multi-dimensional similarity join.
\newblock {\em {VLDB} J.}, 22(4):561--585, 2013.

\bibitem{DBLP:conf/sigmod/SilvaR12}
Yasin~N. Silva and Jason~M. Reed.
\newblock Exploiting mapreduce-based similarity joins.
\newblock In {\em {SIGMOD} Conf. 2012}, pages 693--696, 2012.

\bibitem{DBLP:conf/cloudi/SilvaRT12}
Yasin~N. Silva, Jason~M. Reed, and Lisa~M. Tsosie.
\newblock Mapreduce-based similarity join for metric spaces.
\newblock In {\em Workshop on Cloud Intelligence}, page~3, 2012.

\bibitem{DBLP:conf/icde/BohmK01}
Christian B{\"{o}}hm and Hans{-}Peter Kriegel.
\newblock A cost model and index architecture for the similarity join.
\newblock In {\em ICDE}, pages 411--420, 2001.

\bibitem{DBLP:journals/jda/ParedesR09}
Rodrigo Paredes and Nora Reyes.
\newblock Solving similarity joins and range queries in metric spaces with the
  list of twin clusters.
\newblock {\em J. Discrete Algorithms}, 7(1):18--35, 2009.

\bibitem{DBLP:conf/sigmod/BrinkhoffKS93}
Thomas Brinkhoff, Hans{-}Peter Kriegel, and Bernhard Seeger.
\newblock Efficient processing of spatial joins using r-trees.
\newblock In {\em {SIGMOD} Conf. 1993}, pages 237--246, 1993.

\bibitem{DBLP:conf/vldb/CiacciaPZ97}
Paolo Ciaccia, Marco Patella, and Pavel Zezula.
\newblock M-tree: An efficient access method for similarity search in metric
  spaces.
\newblock In {\em VLDB'97}, pages 426--435, 1997.

\bibitem{DBLP:journals/mta/DohnalGSZ03}
Vlastislav Dohnal, Claudio Gennaro, Pasquale Savino, and Pavel Zezula.
\newblock D-index: Distance searching index for metric data sets.
\newblock {\em Multimedia Tools Appl.}, 21(1):9--33, 2003.

\bibitem{DBLP:conf/dexa/DohnalGZ03}
Vlastislav Dohnal, Claudio Gennaro, and Pavel Zezula.
\newblock Similarity join in metric spaces using ed-index.
\newblock In {\em {DEXA} 2003}, pages 484--493, 2003.

\bibitem{DBLP:conf/sisap/PearsonS14}
Spencer~S. Pearson and Yasin~N. Silva.
\newblock Index-based {R-S} similarity joins.
\newblock In {\em {SISAP}}, pages 106--112, 2014.

\bibitem{DBLP:conf/stoc/IndykM98}
Piotr Indyk and Rajeev Motwani.
\newblock Approximate nearest neighbors: Towards removing the curse of
  dimensionality.
\newblock In {\em Theory of Computing}, pages 604--613, 1998.

\bibitem{DBLP:conf/sigmod/SchneiderD89}
Donovan~A. Schneider and David~J. DeWitt.
\newblock A performance evaluation of four parallel join algorithms in a
  shared-nothing multiprocessor environment.
\newblock In {\em {SIGMOD} Conf. 1989}, pages 110--121, 1989.

\bibitem{DBLP:conf/kdd/WangMP13}
Ye~Wang, Ahmed Metwally, and Srinivasan Parthasarathy.
\newblock Scalable all-pairs similarity search in metric spaces.
\newblock In {\em {SIGKDD}}, pages 829--837, 2013.

\bibitem{DBLP:journals/pvldb/FierABLF18}
Fabian Fier, Nikolaus Augsten, Panagiotis Bouros, Ulf Leser, and
  Johann{-}Christoph Freytag.
\newblock Set similarity joins on mapreduce: An experimental survey.
\newblock {\em {PVLDB}}, 11(10):1110--1122, 2018.

\bibitem{DBLP:conf/waim/LiWU16}
Ye~Li, Jian Wang, and Leong~Hou U.
\newblock Multidimensional similarity join using mapreduce.
\newblock In {\em Web-Age Information Management}, pages 457--468, 2016.

\bibitem{DBLP:conf/sigmod/McCauley018}
Samuel McCauley and Francesco Silvestri.
\newblock Adaptive mapreduce similarity joins.
\newblock In {\em {SIGMOD} Workshop on Algorithms and Systems for MapReduce and
  Beyond}, pages 4:1--4:4, 2018.

\bibitem{DBLP:conf/sigmod/ZhaoRDW16}
Weijie Zhao, Florin Rusu, Bin Dong, and Kesheng Wu.
\newblock Similarity join over array data.
\newblock In {\em {SIGMOD} Conf. 2016}, pages 2007--2022, 2016.

\bibitem{DBLP:conf/btw/BohmNPZ09}
Christian B{\"{o}}hm, Robert Noll, Claudia Plant, and Andrew Zherdin.
\newblock Index-supported similarity join on graphics processors.
\newblock In Johann~Christoph Freytag, Thomas Ruf, Wolfgang Lehner, and
  Gottfried Vossen, editors, {\em Datenbanksysteme in Business, Technologie und
  Web {(BTW} 2009), 13. Fachtagung des GI-Fachbereichs "Datenbanken und
  Informationssysteme" (DBIS), Proceedings, 2.-6. M{\"{a}}rz 2009,
  M{\"{u}}nster, Germany}, volume {P-144} of {\em {LNI}}, pages 57--66. {GI},
  2009.

\bibitem{DBLP:conf/focs/FrigoLPR99}
Matteo Frigo, Charles~E. Leiserson, Harald Prokop, and Sridhar Ramachandran.
\newblock Cache-oblivious algorithms.
\newblock In {\em FOCS 1999}, pages 285--298, 1999.

\bibitem{DBLP:conf/sigmod/HeLLY07}
Bingsheng He, Yinan Li, Qiong Luo, and Dongqing Yang.
\newblock Easedb: a cache-oblivious in-memory query processor.
\newblock In {\em {SIGMOD} Conf. 2007}, pages 1064--1066, 2007.

\bibitem{DBLP:journals/tog/MoonBKCKBNY10}
Bochang Moon, Yongyoung Byun, Tae{-}Joon Kim, Pio Claudio, Hye{-}Sun Kim,
  Yun{-}Ji Ban, Seung~Woo Nam, and Sung{-}Eui Yoon.
\newblock Cache-oblivious ray reordering.
\newblock {\em {ACM} Trans. Graph.}, 29(3), 2010.

\bibitem{DBLP:journals/bmcbi/FerreiraRR14}
Miguel Ferreira, Nuno Roma, and Lu{\'{\i}}s M.~S. Russo.
\newblock Cache-oblivious parallel {SIMD} viterbi decoding for sequence search
  in {HMMER}.
\newblock {\em {BMC} Bioinformatics}, 15:165, 2014.

\bibitem{DBLP:conf/para/BaderM06}
Michael Bader and Christian~E. Mayer.
\newblock Cache oblivious matrix operations using peano curves.
\newblock In {\em PARA Workshop}, pages 521--530, 2006.

\bibitem{DBLP:conf/europar/Bader08}
Michael Bader.
\newblock Exploiting the locality properties of peano curves for parallel
  matrix multiplication.
\newblock In {\em Euro-Par Conference}, pages 801--810, 2008.

\bibitem{loopsjournal}
Christian B{\"{o}}hm, Martin Perdacher, and Claudia Plant.
\newblock A novel hilbert curve for cache-locality preserving loops.
\newblock {\em IEEE Transactions on Big Data}, 2018.

\bibitem{DBLP:conf/bigdataconf/BohmPP16}
Christian B{\"{o}}hm, Martin Perdacher, and Claudia Plant.
\newblock Cache-oblivious loops based on a novel space-filling curve.
\newblock In {\em {IEEE} Big Data}, pages 17--26, 2016.

\bibitem{IEEE:transbigdata/BohmPP18}
Christian B{\"{o}}hm, Martin Perdacher, and Claudia Plant.
\newblock A novel hilbert curve for cache-locality preserving loops.
\newblock {\em IEEE Transactions on Big Data}, pages 1--18, 2018.

\bibitem{DBLP:journals/toms/DongarraCHD90}
Jack Dongarra, Jeremy~Du Croz, Sven Hammarling, and Iain~S. Duff.
\newblock A set of level 3 basic linear algebra subprograms.
\newblock {\em {ACM} Trans. Math. Softw.}, 16(1):1--17, 1990.

\bibitem{DBLP:conf/sigmod/BohmFP08}
Christian B{\"{o}}hm, Christos Faloutsos, and Claudia Plant.
\newblock Outlier-robust clustering using independent components.
\newblock In Jason~Tsong{-}Li Wang, editor, {\em Proceedings of the {ACM}
  {SIGMOD} International Conference on Management of Data, {SIGMOD} 2008,
  Vancouver, BC, Canada, June 10-12, 2008}, pages 185--198. {ACM}, 2008.

\bibitem{DBLP:conf/icdt/BerchtoldBKK01}
Stefan Berchtold, Christian B{\"{o}}hm, Daniel~A. Keim, Florian Krebs, and
  Hans{-}Peter Kriegel.
\newblock On optimizing nearest neighbor queries in high-dimensional data
  spaces.
\newblock In Jan~Van den Bussche and Victor Vianu, editors, {\em Database
  Theory - {ICDT} 2001, 8th International Conference, London, UK, January 4-6,
  2001, Proceedings}, volume 1973 of {\em Lecture Notes in Computer Science},
  pages 435--449. Springer, 2001.

\bibitem{DBLP:conf/icdm/BohmK02}
Christian B{\"{o}}hm and Florian Krebs.
\newblock High performance data mining using the nearest neighbor join.
\newblock In {\em Proceedings of the 2002 {IEEE} International Conference on
  Data Mining {(ICDM} 2002), 9-12 December 2002, Maebashi City, Japan}, pages
  43--50. {IEEE} Computer Society, 2002.

\bibitem{10.1007/BFb0000120}
Rani Siromoney and K.~G. Subramanian.
\newblock Space-filling curves and infinite graphs.
\newblock In Hartmut Ehrig, Manfred Nagl, and Grzegorz Rozenberg, editors, {\em
  Graph-Grammars and Their Application to Computer Science}, pages 380--391,
  Berlin, Heidelberg, 1983. Springer Berlin Heidelberg.

\bibitem{DBLP:conf/icde/BohmOPY07}
Christian B{\"{o}}hm, Beng~Chin Ooi, Claudia Plant, and Ying Yan.
\newblock Efficiently processing continuous k-nn queries on data streams.
\newblock In Rada Chirkova, Asuman Dogac, M.~Tamer {\"{O}}zsu, and Timos~K.
  Sellis, editors, {\em Proceedings of the 23rd International Conference on
  Data Engineering, {ICDE} 2007, The Marmara Hotel, Istanbul, Turkey, April
  15-20, 2007}, pages 156--165. {IEEE} Computer Society, 2007.

\bibitem{DBLP:conf/cikm/BohmBBK00}
Christian B{\"{o}}hm, Bernhard Braunm{\"{u}}ller, Markus~M. Breunig, and
  Hans{-}Peter Kriegel.
\newblock High performance clustering based on the similarity join.
\newblock In {\em Proceedings of the 2000 {ACM} {CIKM} International Conference
  on Information and Knowledge Management, McLean, VA, USA, November 6-11,
  2000}, pages 298--305. {ACM}, 2000.

\bibitem{DBLP:journals/jiis/BohmBKM00}
Christian B{\"{o}}hm, Stefan Berchtold, Hans{-}Peter Kriegel, and Urs Michel.
\newblock Multidimensional index structures in relational databases.
\newblock {\em J. Intell. Inf. Syst.}, 15(1):51--70, 2000.

\bibitem{DBLP:conf/adl/BohmBKS00}
Christian B{\"{o}}hm, Bernhard Braunm{\"{u}}ller, Hans{-}Peter Kriegel, and
  Matthias Schubert.
\newblock Efficient similarity search in digital libraries.
\newblock In {\em Proceedings of {IEEE} Advances in Digital Libraries 2000
  {(ADL} 2000), Washington, DC, USA, May 22-24, 2000}, pages 193--199. {IEEE}
  Computer Society, 2000.

\bibitem{DBLP:conf/edbt/BohmK00}
Christian B{\"{o}}hm and Hans{-}Peter Kriegel.
\newblock Dynamically optimizing high-dimensional index structures.
\newblock In Carlo Zaniolo, Peter~C. Lockemann, Marc~H. Scholl, and Torsten
  Grust, editors, {\em Advances in Database Technology - {EDBT} 2000, 7th
  International Conference on Extending Database Technology, Konstanz, Germany,
  March 27-31, 2000, Proceedings}, volume 1777 of {\em Lecture Notes in
  Computer Science}, pages 36--50. Springer, 2000.

\bibitem{DBLP:journals/sadm/AchtertBDKZ08}
Elke Achtert, Christian B{\"{o}}hm, J{\"{o}}rn David, Peer Kr{\"{o}}ger, and
  Arthur Zimek.
\newblock Global correlation clustering based on the hough transform.
\newblock {\em Statistical Analysis and Data Mining}, 1(3):111--127, 2008.

\bibitem{DBLP:journals/bioinformatics/BaumgartnerBBMWOLR04}
Christian Baumgartner, Christian B{\"{o}}hm, Daniela Baumgartner, G.~Marini,
  Klaus Weinberger, B.~Olgem{\"{o}}ller, B.~Liebl, and A.~A. Roscher.
\newblock Supervised machine learning techniques for the classification of
  metabolic disorders in newborns.
\newblock {\em Bioinform.}, 20(17):2985--2996, 2004.

\bibitem{DBLP:journals/jbi/BaumgartnerBB05}
Christian Baumgartner, Christian B{\"{o}}hm, and Daniela Baumgartner.
\newblock Modelling of classification rules on metabolic patterns including
  machine learning and expert knowledge.
\newblock {\em J. Biomed. Informatics}, 38(2):89--98, 2005.

\bibitem{DBLP:conf/edbt/BohmP08}
Christian B{\"{o}}hm and Claudia Plant.
\newblock {HISSCLU:} a hierarchical density-based method for semi-supervised
  clustering.
\newblock In Alfons Kemper, Patrick Valduriez, Noureddine Mouaddib, Jens
  Teubner, Mokrane Bouzeghoub, Volker Markl, Laurent Amsaleg, and Ioana
  Manolescu, editors, {\em {EDBT} 2008, 11th International Conference on
  Extending Database Technology, Nantes, France, March 25-29, 2008,
  Proceedings}, volume 261 of {\em {ACM} International Conference Proceeding
  Series}, pages 440--451. {ACM}, 2008.

\bibitem{DBLP:conf/kdd/AchtertBKKZ06}
Elke Achtert, Christian B{\"{o}}hm, Hans{-}Peter Kriegel, Peer Kr{\"{o}}ger,
  and Arthur Zimek.
\newblock Deriving quantitative models for correlation clusters.
\newblock In Tina Eliassi{-}Rad, Lyle~H. Ungar, Mark Craven, and Dimitrios
  Gunopulos, editors, {\em Proceedings of the Twelfth {ACM} {SIGKDD}
  International Conference on Knowledge Discovery and Data Mining,
  Philadelphia, PA, USA, August 20-23, 2006}, pages 4--13. {ACM}, 2006.

\bibitem{DBLP:conf/cikm/BohmFOPW09}
Christian B{\"{o}}hm, Frank Fiedler, Annahita Oswald, Claudia Plant, and Bianca
  Wackersreuther.
\newblock Probabilistic skyline queries.
\newblock In David~Wai{-}Lok Cheung, Il{-}Yeol Song, Wesley~W. Chu, Xiaohua Hu,
  and Jimmy~J. Lin, editors, {\em Proceedings of the 18th {ACM} Conference on
  Information and Knowledge Management, {CIKM} 2009, Hong Kong, China, November
  2-6, 2009}, pages 651--660. {ACM}, 2009.

\bibitem{DBLP:conf/ssdbm/BohmPS06}
Christian B{\"{o}}hm, Alexey Pryakhin, and Matthias Schubert.
\newblock Probabilistic ranking queries on gaussians.
\newblock In {\em 18th International Conference on Scientific and Statistical
  Database Management, {SSDBM} 2006, 3-5 July 2006, Vienna, Austria,
  Proceedings}, pages 169--178. {IEEE} Computer Society, 2006.

\bibitem{DBLP:conf/kdd/BohmHMP09}
Christian B{\"{o}}hm, Katrin Haegler, Nikola~S. M{\"{u}}ller, and Claudia
  Plant.
\newblock Coco: coding cost for parameter-free outlier detection.
\newblock In John F.~Elder IV, Fran{\c{c}}oise Fogelman{-}Souli{\'{e}},
  Peter~A. Flach, and Mohammed~Javeed Zaki, editors, {\em Proceedings of the
  15th {ACM} {SIGKDD} International Conference on Knowledge Discovery and Data
  Mining, Paris, France, June 28 - July 1, 2009}, pages 149--158. {ACM}, 2009.

\bibitem{DBLP:conf/dawak/BerchtoldBKKX00}
Stefan Berchtold, Christian B{\"{o}}hm, Daniel~A. Keim, Hans{-}Peter Kriegel,
  and Xiaowei Xu.
\newblock Optimal multidimensional query processing using tree striping.
\newblock In Yahiko Kambayashi, Mukesh~K. Mohania, and A~Min Tjoa, editors,
  {\em Data Warehousing and Knowledge Discovery, Second International
  Conference, DaWaK 2000, London, UK, September 4-6, 2000, Proceedings}, volume
  1874 of {\em Lecture Notes in Computer Science}, pages 244--257. Springer,
  2000.

\bibitem{DBLP:conf/sdm/AchtertBDKZ08}
Elke Achtert, Christian B{\"{o}}hm, J{\"{o}}rn David, Peer Kr{\"{o}}ger, and
  Arthur Zimek.
\newblock Robust clustering in arbitrarily oriented subspaces.
\newblock In {\em Proceedings of the {SIAM} International Conference on Data
  Mining, {SDM} 2008, April 24-26, 2008, Atlanta, Georgia, {USA}}, pages
  763--774. {SIAM}, 2008.

\bibitem{DBLP:journals/jdi/BaumgartnerGBF05}
Christian Baumgartner, Kurt Gautsch, Christian B{\"{o}}hm, and Stephan Felber.
\newblock Functional cluster analysis of {CT} perfusion maps: {A} new tool for
  diagnosis of acute stroke?
\newblock {\em J. Digital Imaging}, 18(3):219--226, 2005.

\bibitem{DBLP:journals/kais/MaiHFPB15}
Son~T. Mai, Xiao He, Jing Feng, Claudia Plant, and Christian B{\"{o}}hm.
\newblock Anytime density-based clustering of complex data.
\newblock {\em Knowl. Inf. Syst.}, 45(2):319--355, 2015.

\bibitem{DBLP:journals/tlsdkcs/BohmNPWZ09}
Christian B{\"{o}}hm, Robert Noll, Claudia Plant, Bianca Wackersreuther, and
  Andrew Zherdin.
\newblock Data mining using graphics processing units.
\newblock {\em Trans. Large Scale Data Knowl. Centered Syst.}, 1:63--90, 2009.

\bibitem{DBLP:conf/dexa/BohmK03}
Christian B{\"{o}}hm and Florian Krebs.
\newblock Supporting {KDD} applications by the k-nearest neighbor join.
\newblock In Vladim{\'{\i}}r Mar{\'{\i}}k, Werner Retschitzegger, and Olga
  Step{\'{a}}nkov{\'{a}}, editors, {\em Database and Expert Systems
  Applications, 14th International Conference, {DEXA} 2003, Prague, Czech
  Republic, September 1-5, 2003, Proceedings}, volume 2736 of {\em Lecture
  Notes in Computer Science}, pages 504--516. Springer, 2003.

\bibitem{DBLP:conf/icde/BohmGKPS07}
Christian B{\"{o}}hm, Michael Gruber, Peter Kunath, Alexey Pryakhin, and
  Matthias Schubert.
\newblock Prover: Probabilistic video retrieval using the gauss-tree.
\newblock In Rada Chirkova, Asuman Dogac, M.~Tamer {\"{O}}zsu, and Timos~K.
  Sellis, editors, {\em Proceedings of the 23rd International Conference on
  Data Engineering, {ICDE} 2007, The Marmara Hotel, Istanbul, Turkey, April
  15-20, 2007}, pages 1521--1522. {IEEE} Computer Society, 2007.

\bibitem{DBLP:conf/miccai/DyrbaEWKPOMPBFFHKHKT12}
Martin Dyrba, Michael Ewers, Martin Wegrzyn, Ingo Kilimann, Claudia Plant,
  Annahita Oswald, Thomas Meindl, Michela Pievani, Arun L.~W. Bokde, Andreas
  Fellgiebel, Massimo Filippi, Harald Hampel, Stefan Kl{\"{o}}ppel, Karlheinz
  Hauenstein, Thomas Kirste, and Stefan~J. Teipel.
\newblock Combining {DTI} and {MRI} for the automated detection of alzheimer's
  disease using a large european multicenter dataset.
\newblock In Pew{-}Thian Yap, Tianming Liu, Dinggang Shen, Carl{-}Fredrik
  Westin, and Li~Shen, editors, {\em Multimodal Brain Image Analysis - Second
  International Workshop, {MBIA} 2012, Held in Conjunction with {MICCAI} 2012,
  Nice, France, October 1-5, 2012. Proceedings}, volume 7509 of {\em Lecture
  Notes in Computer Science}, pages 18--28. Springer, 2012.

\bibitem{DBLP:conf/kdd/PlantB11}
Claudia Plant and Christian B{\"{o}}hm.
\newblock {INCONCO:} interpretable clustering of numerical and categorical
  objects.
\newblock In Chid Apt{\'{e}}, Joydeep Ghosh, and Padhraic Smyth, editors, {\em
  Proceedings of the 17th {ACM} {SIGKDD} International Conference on Knowledge
  Discovery and Data Mining, San Diego, CA, USA, August 21-24, 2011}, pages
  1127--1135. {ACM}, 2011.

\bibitem{DBLP:journals/bioinformatics/PlantBTB06}
Claudia Plant, Christian B{\"{o}}hm, Bernhard Tilg, and Christian Baumgartner.
\newblock Enhancing instance-based classification with local density: a new
  algorithm for classifying unbalanced biomedical data.
\newblock {\em Bioinform.}, 22(8):981--988, 2006.

\bibitem{DBLP:conf/icdm/ShaoPYB11}
Junming Shao, Claudia Plant, Qinli Yang, and Christian B{\"{o}}hm.
\newblock Detection of arbitrarily oriented synchronized clusters in
  high-dimensional data.
\newblock In Diane~J. Cook, Jian Pei, Wei Wang, Osmar~R. Za{\"{\i}}ane, and
  Xindong Wu, editors, {\em 11th {IEEE} International Conference on Data
  Mining, {ICDM} 2011, Vancouver, BC, Canada, December 11-14, 2011}, pages
  607--616. {IEEE} Computer Society, 2011.

\bibitem{DBLP:conf/icdm/PlantWZ09}
Claudia Plant, Afra~M. Wohlschl{\"{a}}ger, and Andrew Zherdin.
\newblock Interaction-based clustering of multivariate time series.
\newblock In Wei Wang, Hillol Kargupta, Sanjay Ranka, Philip~S. Yu, and Xindong
  Wu, editors, {\em {ICDM} 2009, The Ninth {IEEE} International Conference on
  Data Mining, Miami, Florida, USA, 6-9 December 2009}, pages 914--919. {IEEE}
  Computer Society, 2009.

\bibitem{alzheimer}
Junming Shao, Nicholas Myers, Qinli Yang, Jing Feng, Claudia Plant, Christian
  Böhm, Hans Förstl, Alexander Kurz, Claus Zimmer, Chun Meng, Valentin Riedl,
  Afra Wohlschläger, and Christian Sorg.
\newblock Prediction of alzheimer's disease using individual structural
  connectivity networks.
\newblock {\em Neurobiology of aging}, 33(12):2756--2765.

\bibitem{DBLP:journals/tkdd/BohmFPP07}
Christian B{\"{o}}hm, Christos Faloutsos, Jia{-}Yu Pan, and Claudia Plant.
\newblock {RIC:} parameter-free noise-robust clustering.
\newblock {\em {ACM} Trans. Knowl. Discov. Data}, 1(3):10, 2007.

\bibitem{DBLP:conf/pakdd/BohmGOPPW10}
Christian B{\"{o}}hm, Sebastian Goebl, Annahita Oswald, Claudia Plant, Michael
  Plavinski, and Bianca Wackersreuther.
\newblock Integrative parameter-free clustering of data with mixed type
  attributes.
\newblock In Mohammed~Javeed Zaki, Jeffrey~Xu Yu, Balaraman Ravindran, and
  Vikram Pudi, editors, {\em Advances in Knowledge Discovery and Data Mining,
  14th Pacific-Asia Conference, {PAKDD} 2010, Hyderabad, India, June 21-24,
  2010. Proceedings. Part {I}}, volume 6118 of {\em Lecture Notes in Computer
  Science}, pages 38--47. Springer, 2010.

\bibitem{DBLP:conf/icdm/MaiGP12}
Son~T. Mai, Sebastian Goebl, and Claudia Plant.
\newblock A similarity model and segmentation algorithm for white matter fiber
  tracts.
\newblock In Mohammed~Javeed Zaki, Arno Siebes, Jeffrey~Xu Yu, Bart Goethals,
  Geoffrey~I. Webb, and Xindong Wu, editors, {\em 12th {IEEE} International
  Conference on Data Mining, {ICDM} 2012, Brussels, Belgium, December 10-13,
  2012}, pages 1014--1019. {IEEE} Computer Society, 2012.

\bibitem{DBLP:journals/kais/ShaoWYPB17}
Junming Shao, Xinzuo Wang, Qinli Yang, Claudia Plant, and Christian B{\"{o}}hm.
\newblock Synchronization-based scalable subspace clustering of
  high-dimensional data.
\newblock {\em Knowl. Inf. Syst.}, 52(1):83--111, 2017.

\bibitem{DBLP:conf/kdd/YeGPB16}
Wei Ye, Sebastian Goebl, Claudia Plant, and Christian B{\"{o}}hm.
\newblock {FUSE:} full spectral clustering.
\newblock In Balaji Krishnapuram, Mohak Shah, Alexander~J. Smola, Charu~C.
  Aggarwal, Dou Shen, and Rajeev Rastogi, editors, {\em Proceedings of the 22nd
  {ACM} {SIGKDD} International Conference on Knowledge Discovery and Data
  Mining, San Francisco, CA, USA, August 13-17, 2016}, pages 1985--1994. {ACM},
  2016.

\bibitem{DBLP:conf/kdd/FengHKBP12}
Jing Feng, Xiao He, Bettina Konte, Christian B{\"{o}}hm, and Claudia Plant.
\newblock Summarization-based mining bipartite graphs.
\newblock In Qiang Yang, Deepak Agarwal, and Jian Pei, editors, {\em The 18th
  {ACM} {SIGKDD} International Conference on Knowledge Discovery and Data
  Mining, {KDD} '12, Beijing, China, August 12-16, 2012}, pages 1249--1257.
  {ACM}, 2012.

\bibitem{DBLP:journals/envsoft/YangSSBP12}
Qinli Yang, Junming Shao, Miklas Scholz, Christian B{\"{o}}hm, and Claudia
  Plant.
\newblock Multi-label classification models for sustainable flood retention
  basins.
\newblock {\em Environ. Model. Softw.}, 32:27--36, 2012.

\bibitem{DBLP:conf/icdm/GoeblHPB14}
Sebastian Goebl, Xiao He, Claudia Plant, and Christian B{\"{o}}hm.
\newblock Finding the optimal subspace for clustering.
\newblock In Ravi Kumar, Hannu Toivonen, Jian Pei, Joshua~Zhexue Huang, and
  Xindong Wu, editors, {\em 2014 {IEEE} International Conference on Data
  Mining, {ICDM} 2014, Shenzhen, China, December 14-17, 2014}, pages 130--139.
  {IEEE} Computer Society, 2014.

\bibitem{DBLP:conf/kdd/AltinigneliPB13}
Muzaffer~Can Altinigneli, Claudia Plant, and Christian B{\"{o}}hm.
\newblock Massively parallel expectation maximization using graphics processing
  units.
\newblock In Inderjit~S. Dhillon, Yehuda Koren, Rayid Ghani, Ted~E. Senator,
  Paul Bradley, Rajesh Parekh, Jingrui He, Robert~L. Grossman, and Ramasamy
  Uthurusamy, editors, {\em The 19th {ACM} {SIGKDD} International Conference on
  Knowledge Discovery and Data Mining, {KDD} 2013, Chicago, IL, USA, August
  11-14, 2013}, pages 838--846. {ACM}, 2013.

\bibitem{DBLP:conf/icdm/YeMHP16}
Wei Ye, Samuel Maurus, Nina Hubig, and Claudia Plant.
\newblock Generalized independent subspace clustering.
\newblock In Francesco Bonchi, Josep Domingo{-}Ferrer, Ricardo Baeza{-}Yates,
  Zhi{-}Hua Zhou, and Xindong Wu, editors, {\em {IEEE} 16th International
  Conference on Data Mining, {ICDM} 2016, December 12-15, 2016, Barcelona,
  Spain}, pages 569--578. {IEEE} Computer Society, 2016.

\bibitem{DBLP:conf/kdd/Plant12}
Claudia Plant.
\newblock Dependency clustering across measurement scales.
\newblock In Qiang Yang, Deepak Agarwal, and Jian Pei, editors, {\em The 18th
  {ACM} {SIGKDD} International Conference on Knowledge Discovery and Data
  Mining, {KDD} '12, Beijing, China, August 12-16, 2012}, pages 361--369.
  {ACM}, 2012.

\bibitem{DBLP:journals/kais/BohmK04}
Christian B{\"{o}}hm and Florian Krebs.
\newblock The \emph{k}-nearest neighbour join: Turbo charging the {KDD}
  process.
\newblock {\em Knowl. Inf. Syst.}, 6(6):728--749, 2004.

\bibitem{DBLP:conf/mlg/WackersreutherW10}
Bianca Wackersreuther, Peter Wackersreuther, Annahita Oswald, Christian
  B{\"{o}}hm, and Karsten~M. Borgwardt.
\newblock Frequent subgraph discovery in dynamic networks.
\newblock In {\em MLG@KDD}, pages 155--162. {ACM}, 2010.

\bibitem{DBLP:conf/ssdbm/AchtertBKKZ07}
Elke Achtert, Christian B{\"{o}}hm, Hans{-}Peter Kriegel, Peer Kr{\"{o}}ger,
  and Arthur Zimek.
\newblock On exploring complex relationships of correlation clusters.
\newblock In {\em {SSDBM}}, page~7. {IEEE} Computer Society, 2007.

\bibitem{DBLP:journals/tkde/ShaoHBYP13}
Junming Shao, Xiao He, Christian B{\"{o}}hm, Qinli Yang, and Claudia Plant.
\newblock Synchronization-inspired partitioning and hierarchical clustering.
\newblock {\em {IEEE} Trans. Knowl. Data Eng.}, 25(4):893--905, 2013.

\bibitem{DBLP:conf/pkdd/AchtertBKKMZ06}
Elke Achtert, Christian B{\"{o}}hm, Hans{-}Peter Kriegel, Peer Kr{\"{o}}ger,
  Ina M{\"{u}}ller{-}Gorman, and Arthur Zimek.
\newblock Finding hierarchies of subspace clusters.
\newblock In {\em {PKDD}}, volume 4213 of {\em Lecture Notes in Computer
  Science}, pages 446--453. Springer, 2006.

\bibitem{DBLP:conf/ssdbm/AchtertBKZ06}
Elke Achtert, Christian B{\"{o}}hm, Peer Kr{\"{o}}ger, and Arthur Zimek.
\newblock Mining hierarchies of correlation clusters.
\newblock In {\em {SSDBM}}, pages 119--128. {IEEE} Computer Society, 2006.

\bibitem{patent}
Stefan Berchtold, Christian B{\"{o}}hm, and Hans-Peter Kriegel.
\newblock High-dimensional index structure.
\newblock In {\em US Patent {6,154,746}}, 2000.

\bibitem{DBLP:conf/kdd/MaurusP16}
Samuel Maurus and Claudia Plant.
\newblock Skinny-dip: Clustering in a sea of noise.
\newblock In {\em {KDD}}, pages 1055--1064. {ACM}, 2016.

\bibitem{DBLP:conf/icdm/MaiHHPB13}
Son~T. Mai, Xiao He, Nina Hubig, Claudia Plant, and Christian B{\"{o}}hm.
\newblock Active density-based clustering.
\newblock In {\em {ICDM}}, pages 508--517. {IEEE} Computer Society, 2013.

\bibitem{water}
Qinli Yang, Junming Shao, Miklas Scholz, and Claudia Plant.
\newblock Feature selection methods for characterizing and classifying adaptive
  sustainable flood retention basins.
\newblock {\em Water Research}, 45(3):993--1004, 2011.

\bibitem{DBLP:journals/isci/ShaoTGYPA19}
Junming Shao, Yue Tan, Lianli Gao, Qinli Yang, Claudia Plant, and Ira Assent.
\newblock Synchronization-based clustering on evolving data stream.
\newblock {\em Inf. Sci.}, 501:573--587, 2019.

\bibitem{DBLP:conf/sdm/BohmPP17}
Christian B{\"{o}}hm, Martin Perdacher, and Claudia Plant.
\newblock Multi-core k-means.
\newblock In {\em SDM}, pages 273--281, 2017.

\bibitem{DBLP:conf/icdm/GoeblTBP16}
Sebastian Goebl, Annika Tonch, Christian B{\"{o}}hm, and Claudia Plant.
\newblock Megs: Partitioning meaningful subgraph structures using minimum
  description length.
\newblock In {\em {ICDM}}, pages 889--894. {IEEE} Computer Society, 2016.

\bibitem{DBLP:conf/kdd/MautzYPB17}
Dominik Mautz, Wei Ye, Claudia Plant, and Christian B{\"{o}}hm.
\newblock Towards an optimal subspace for k-means.
\newblock In {\em {KDD}}, pages 365--373. {ACM}, 2017.

\bibitem{Bially1969SpacefillingCT}
Theodore Bially.
\newblock Space-filling curves: Their generation and their application to
  bandwidth reduction.
\newblock {\em IEEE Trans. Information Theory}, 15:658--664, 1969.

\bibitem{Prusinkiewicz:1986:GAL:16564.16608}
P~Prusinkiewicz.
\newblock Graphical applications of l-systems.
\newblock In {\em Proceedings on Graphics Interface '86/Vision Interface '86},
  pages 247--253, Toronto, Ont., Canada, Canada, 1986. Canadian Information
  Processing Society.

\end{thebibliography}
\end{document}